\documentclass[lettersize,journal]{IEEEtran}
\usepackage{amsmath,amsfonts}
\usepackage{algorithmic}
\usepackage{algorithm}
\usepackage{array}
\usepackage[caption=false,font=normalsize,labelfont=sf,textfont=sf]{subfig}
\usepackage{textcomp}
\usepackage{stfloats}
\usepackage{url}
\usepackage{verbatim}
\usepackage{graphicx}
\usepackage{cite}
\usepackage{hyperref}
\hypersetup{hidelinks}
\hypersetup{colorlinks}
\usepackage{multirow}
\usepackage{enumerate}
\usepackage{booktabs}
\usepackage{makecell}
\usepackage{ulem}
\usepackage{float}

\begin{document}

\title{Adaptive Split-Fusion Transformer}

\author{Zixuan Su, Hao Zhang, Jingjing Chen,~\IEEEmembership{Member,~IEEE}, Lei Pang, Chong-Wah Ngo,~\IEEEmembership{Member,~IEEE}\\ and Yu-Gang Jiang,~\IEEEmembership{Senior Member,~IEEE}
\thanks{Zixuan Su, Jingjing Chen and Yu-Gang Jiang are with School of Computer Science, Fudan University, Shanghai, China (E-mail: zxsu21@m.fudan.edu.cn, \{chenjingjing,ygj\}@fudan.edu.cn). Hao Zhang and Chong-Wah Ngo are with Singapore Management University, Singapore (E-mail: \{hzhang,cwngo\}@smu.edu.sg). Lei Pang is with City University of Hong Kong, Hong Kong, China (E-mail: cactuslei@gmail.com).}
\thanks{The Corresponding author is Jingjing Chen.}
}



\maketitle

\begin{figure*}
\centering
\includegraphics[height=10cm]{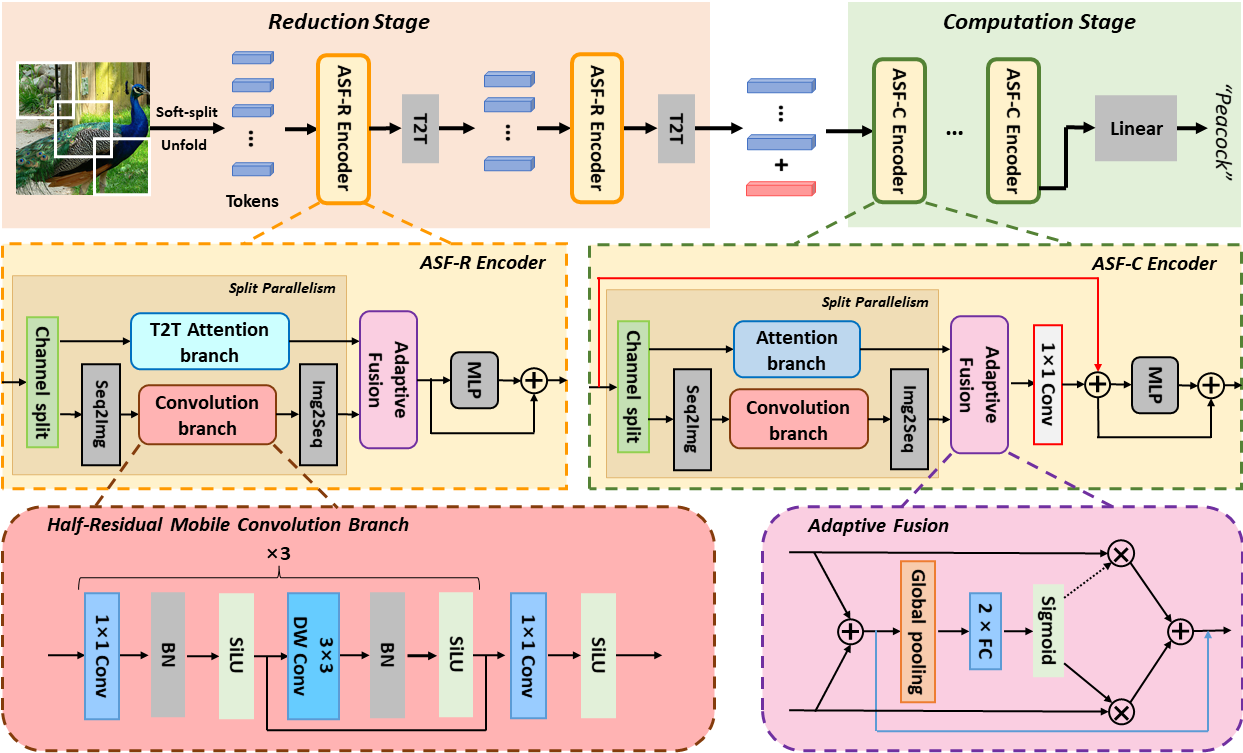}
\caption{An overview of the ASF-former. The encoders in a reduction and computation stage are separately denoted as ASR-R/C. Both types include Split Parallelism and Adaptive Fusion parts, except that the ASF-R adopts T2T attention for down-sampling token dimension. As shortcut and Conv $1\times1$({\color{red}red} line) is incompatible with the T2T attention \cite{yuan2021tokens}, they are removed in the reduction stage. (This figure is best viewed in color)
}
\label{fig:framework}
\end{figure*}

\begin{abstract}
Neural networks for visual content understanding have recently evolved from convolutional ones to transformers. The prior (CNN) relies on small-windowed kernels to capture the regional clues, demonstrating solid local expressiveness. On the contrary, the latter (transformer) establishes long-range global connections between localities for holistic learning. Inspired by this complementary nature, there is a growing interest in designing hybrid models which utilize both techniques. 
Current hybrids merely replace convolutions as simple approximations of linear projection or juxtapose a convolution branch with attention without considering the importance of local/global modeling. To tackle this, we propose a new hybrid named \textit{Adaptive Split-Fusion Transformer} (ASF-former) that treats convolutional and attention branches differently with adaptive weights. Specifically, an ASF-former encoder equally splits feature channels into half to fit dual-path inputs. Then, the outputs of the dual-path are fused with weights calculated from visual 
cues. We also design a compact convolutional path from a concern of efficiency. Extensive experiments on standard benchmarks, such as ImageNet-1K, CIFAR-10, and CIFAR-100, show that our ASF-former outperforms its CNN, transformer, and hybrid counterparts in terms of accuracy (83.9\% on ImageNet-1K), under similar conditions (12.9G MACs/56.7M Params, without large-scale pre-training). The code is available at: \url{https://github.com/szx503045266/ASF-former}.
\end{abstract}


\section{Introduction}

Neural networks for learning visual representations have recently split into directions of conventional convolutional neural networks (i.e., CNN) and emerging transformers. CNN used to be the de-facto standard network and was good at modeling localities. On the contrary, the transformer learns holistic features by building pair-wise relations, thus demonstrating strong global expressiveness. Pilot visual transformers, including ViT \cite{dosovitskiy2021an}, T2T-ViT \cite{yuan2021tokens}, deliberately avoid convolutions and only rely on the self-attention. Though achieving good accuracy, they pay extra computations as a price for bypassing efficient convolution operators. 

Since convolutions and self-attention are complementary when viewed from perspectives like local-global modeling and high-low efficiency, it is natural to study hybrid networks to ensure each part serves its best. Existing hybrids usually combine these two parts in a cascade or parallel manner. For a cascade hybrid, researchers usually re-implement linear projections in vanilla transformers with convolutional approximations. For example, token-embedding \cite{Chen2021VisformerTV,Hassani2021EscapingTB,Mehta2021MobileViTLG,Wu2021CvTIC,Yan2021ConTNetWN,Yang2021FocalSF,Yuan2021IncorporatingCD} and linear-projections \cite{Chen2021VisformerTV,He2021PruningSI,Mehta2021MobileViTLG,Wu2021CvTIC,Yan2021ConTNetWN,Yuan2021IncorporatingCD} in attentions/MLPs are commonly replaced by convolutions. These cascade works share a common principle of minimal change. As for parallel hybrids, an extra convolutional branch is inserted on par with the attention in a dual-branch (or dual-path) manner \cite{Chen2021MobileFormerBM,Pan2021OnTI,Peng2021ConformerLF,Xu2021ViTAEVT}. This strategy enables learning local/global visual contexts independently and is beneficial for analyzing the effectiveness of each path.

However, most current hybrid models treat local and global contexts equally, which conflicts with the real-world scenario that the importance of local/global cues varies according to the image visuals and layer depth. For example, tiny objects prefer local evidence, whereas landscapes bias global views in the recognition process. Besides, layer with different depths also shows their biases in learning local/global contexts, as mentioned in \cite{Pan2021OnTI}.

To tackle this, we propose a novel parallel hybrid named {\textit{Adaptive Split-Fusion Transformer}} (ASF-former), which adopts an adaptive gating strategy to select convolution/attention paths according to global visual cues. Its encoder contains two parts: \textit{Efficient Split Parallelism with HMCB} and \textit{Adaptive Fusion} as shown in Figure~(\ref{fig:framework}).

\textbf{Efficient Split Parallelism with HMCB} differs from the existing parallel hybrid models in two aspects. Firstly, we comprehensively and carefully craft an efficient convolution path named \textit{Half-Residual Mobile Convolutional Branch} (HMCB). Compared with existing counterparts, the HMCB demonstrates stronger local capability with fewer computations. Secondly, we split inherent feature channels of pure transformers into half and separately feed sub-features into the Conv/attention branch. Thereby, we could keep the same feature dimension as the original backbone. With these, the Split Parallelism shares a similar complexity as single-path (convolution or attention) models.

\textbf{Adaptive Fusion} module intakes outputs from convolution and attention branches and weighs them with adaptive scalars. Specifically, visual features from both paths are processed by sequential layers, including global pooling, fully-connected layer, and Sigmoid activation, to generate weighting scalars. These scalars are used to weigh features from each branch (Fig. (\ref{fig:framework})). We also add an extra skip connection to alleviate gradient vanishing in backpropagation. We experimentally verify that the new adaptive fusion could effectively and efficiently select convolution/attention branches according to visual contents. We briefly summarize our contributions below.

\hspace*{\fill} \\

\begin{itemize}
    \item \textit{Efficient Split Parallelism with HMCB}. We introduce a new Half-Residual Mobile Convolutional Branch, which complements the attention feature with high cost-effectiveness. 
    \item \textit{Adaptive Fusion}. We first introduce adaptive fusion to combine outputs from convolution and attention branches. We could adjust the importance of local/global modeling with adaptive weights according to visual contents.
    \item \textit{A new hybrid ASF-former}. We build a new CNN-transformer hybrid with efficient convolution layers (HMCB) and an adaptive fusion strategy. Experiments on standard benchmarks, such as ImageNet-1K and downstream datasets, show that the ASF-former could achieve SOTA performance (83.9\% on ImageNet-1K) under similar conditions (12.9G MACs / 56.7M Params). 
\end{itemize}

\section{Related Work}

\textbf{Convolutional Neural Network.} Since the birth of AlexNet~\cite{Krizhevsky2012ImageNetCW}, convolutional neural network (CNN) was the de-facto standard model on different vision topics, such as image recognition~\cite{Simonyan2015VeryDC,Szegedy2015GoingDW,Szegedy2016RethinkingTI,He2016DeepRL,He2016IdentityMI,Huang2017DenselyCC}, object detection~\cite{Girshick2015FastR,Ren2015FasterRT,Redmon2016YouOL,Liu2016SSDSS,Lin2017FocalLF,Lin2017FeaturePN,Cai2018CascadeRD}, instance/semantic segmentation~\cite{He2017MaskR,Chen2017RethinkingAC,Chen2018DeepLabSI} and video recognition~\cite{Feichtenhofer2016ConvolutionalTN,Qiu2017LearningSR,Carreira2017QuoVA,Zhou2018TemporalRR,Wang2018NonlocalNN,Feichtenhofer2019SlowFastNF,Lin2019TSMTS}.

To apply CNN to edge devices, researchers spare many efforts on balancing complexity  and capacity. For example, MobileNet~\cite{howard2017mobilenets} factorized a standard convolution with depthwise separable convolution. Moreover, MobileNetV2~\cite{Sandler2018MobileNetV2IR} introduced an extra inverted residual design and expanded the usage of depthwise convolution into more layers (e.g., intermedia expansion layer). ShuffleNet~\cite{Zhang2018ShuffleNetAE} combines group convolution, channel shuffle, and depthwise convolution to build an efficient CNN unit. EfficientNet~\cite{Tan2019EfficientNetRM} follows an empirical scaling rule to search and adjust hyperparameters of network structure, such as ``\textit{dimension, depth, width, resolution}'', to achieve a high cost-effective.

Some researchers proposed plug-and-play modules introducing a few extra parameters to recalibrate CNN features. For example, the SE-Net~\cite{Hu2018SqueezeandExcitationN} introduced a new ``Squeeze-and-Excitation'' block on top of a plain 2D-CNN backbone. The block could adaptively adjust feature channels with global contexts. Furthermore, GE-Net~\cite{Hu2018GatherExciteEF} studied more choices of ``Gathering'' contexts information than SE-Net. Hao et al. \cite{hao2022group} extended contextual calibrating for videos and proposed GC-Net that utilizes spatial-temporal contexts. SKNet~\cite{Li2019SelectiveKN} selectively fuse two convolution branches with different kernel sizes to adaptively adjust the receptive field size. ResNeSt~\cite{Zhang2020ResNeStSN} further combined
the channel-wise attention with a multi-path strategy. This design would build cross-feature interactions and benefit learning diverse representations. 

\textbf{Vision Transformer.} Transformer receives extensive interest in various multimedia and vision tasks, such as Video Recognition \cite{alfasly2022effective}, Image Understanding \cite{EAPT, YDTR}, Re-Identification \cite{Li2021ExploitingMP, Zhao2022SpatialET,zhou2022moving}, Visual Inpainting \cite{FT-TDR}, Pose Estimation \cite{Li2022ExploitingTC} tasks since the birth of ViT \cite{dosovitskiy2021an}, which validates the feasibility of replacing CNNs with pure transformers with large-scale pre-training. 

Though achieving impressive accuracy, the ViT \cite{dosovitskiy2021an} suffers from a high computational cost. The cost is caused by densely calculating the pair-wise distance between visual tokens in each attention module. To balance computations and classification accuracy, researchers spare more effort on developing new transformers, including DeiT \cite{Touvron2021TrainingDI}, T2T-ViT \cite{yuan2021tokens}, TNT \cite{tnt}, PVT \cite{wang2021pyramid}, Swin \cite{liu2021Swin} than before. Specifically, DeiT \cite{Touvron2021TrainingDI} adopted convnet as a teacher and trained transformer under a teacher-student strategy. It relies on the distillation token to introduce locality into a transformer, thus lowering the requirement for large-scale training data.
T2T-ViT \cite{yuan2021tokens} focused on shrinking token-length. It designs the T2T module to down-sampling tokens via concatenating features of local neighboring pixels.
TNT~\cite{tnt} altered the granularity of patch dividing by further splitting local patches into sub-patches. The inner attention will be calculated within each patch before outer global attention.
In order to be ported to various downstream tasks, PVT~\cite{wang2021pyramid} introduced a progressive shrinking pyramid structure and a spatial-reduction attention for efficiently learning multi-scale and high-resolution features.
For further parameter and computation efficiency, Swin Transformer \cite{liu2021Swin} utilized shifted window to split the feature map and performed self-attention within each local window. 
These models are pure convolutional-free transformers, thus lacking local capacity and convolution's efficiency strength.

\textbf{Hybrid Transformer.} Attracted by the complementary nature of CNN and Attentions, more and more efforts are devoted to developing hybrid transformers. Existing hybrids can be separated into two groups.
The first is cascade hybrid which minimally modify the original transformer model by re-implementing the token-embedding \cite{Chen2021VisformerTV,Hassani2021EscapingTB,Mehta2021MobileViTLG,Wu2021CvTIC,Yan2021ConTNetWN,Yang2021FocalSF,Yuan2021IncorporatingCD} and the linear projections \cite{Chen2021VisformerTV,He2021PruningSI,Mehta2021MobileViTLG,Wu2021CvTIC,Yan2021ConTNetWN,Yuan2021IncorporatingCD} in Attentions/MLPs with convolution operators.
The second is parallel hybrid which juxtaposes an extra convolutional branch on par with the attention \cite{Chen2021MobileFormerBM,Pan2021OnTI,Peng2021ConformerLF,Xu2021ViTAEVT}. For example, Conformer \cite{Peng2021ConformerLF} designed the Feature Coupling Unit (FCU) for transmitting features from one path to another. For acquiring inductive bias from convolution, ViTAE \cite{Xu2021ViTAEVT} built the parallel structure in each block and designed the pyramid reduction module with dilated convolution. These methods treat convolution and attention paths equally. 
ACmix \cite{Pan2021OnTI} instead set two learnable weights for measuring the importance of two paths, but the weights only vary with network depth, failing to be adjusted according to visual contexts.

\section{Our Method}
An overview of ASF-former is shown in Figure~(\ref{fig:framework}). Similar to \cite{Xu2021ViTAEVT,yuan2021tokens}, it contains a total of $L=L_1+L_2$ encoders, where $L_1$/$L_2$ encoders reside in \textit{reduction} or \textit{computation} stages. As in \cite{yuan2021tokens}, the two stages differentiate in whether adopting the T2T for shrinking token-length and T2T attentions for reducing computations. To distinguish, we separately denote encoders in the two stages as the ASF-R and ASF-C. We present a detailed pipeline of ASF-former below.

An image $\boldsymbol{I} \in {\mathbb{R}}^{H \times W \times 3}$ is first soft-split into patches. Each patch shares an identical shape of $k \times k$ with overlap $o$ and padding $p$. These patches are unfolded into a sequence of tokens $\boldsymbol{T}_0 \in {\mathbb{R}}^{N_0 \times D_0}$, where $D_0=3k^2$, and token-length is:
\begin{equation}
    N_0 = \biggl\lfloor{\frac{H+2p-k}{k-o}+1}\biggr\rfloor 
    \times \biggl\lfloor{\frac{W+2p-k}{k-o}+1}\biggr\rfloor.
\label{patch_number}
\end{equation}
Tokens $\boldsymbol{T}_0$ go through the two stages, including the reduction and computation stages for representation learning.  

\textbf{Reduction stage} contains $L_1$ replicated ASF-R + T2T pairs, where the prior and the latter module separately serve for feature learning and down-sampling. Denote tokens from the $i$-th pair as $\boldsymbol{T}_{i}$ $\in \mathbb{R}^{N_{i} \times D_{i}}$ or $\boldsymbol{\widetilde{T}}_{i}$ $\in \mathbb{R}^{N_{i} \times D'}$. The token-length $N_i$ and dimension $D_i$ would reduce and increase according to the depth $i \in [1,2,\cdots, L_1]$, due to the T2T operation, while the ASF-R encoder would decrease the token dimension to $D'$. A math process of the $i$-th pair is shown as:
\begin{align}
    &\boldsymbol{\widetilde{T}}_{i-1}=f_{ar}(\boldsymbol{T}_{i-1})\label{eq:f1}\\
    &\boldsymbol{{T}}_{i}=f_{t2t}(\boldsymbol{\widetilde{T}}_{i-1})\label{eq:f2}
\end{align}
where $f_{ar}\left(\cdot\right)$ and $f_{t2t}\left(\cdot\right)$ denotes the ASR-R and T2T modules.

Output $\boldsymbol{T}_{out} \in \mathbb{R}^{N_{L_1} \times D}$ of reduction stage is obtained by linear-projecting $\boldsymbol{T}_{L_1}$ to a fixed $D$-dimensional space.
\begin{align}
    &\boldsymbol{T}_{out}=Linear\left(\boldsymbol{T}_{L_1}\right)
\end{align}

\textbf{Computation stage} contains $L_2$ identical ASF-C encoders, without changing token-length. Same as the ViT \cite{dosovitskiy2021an}, an extra [{\tt CLASS}] token $\boldsymbol{C}_0 \in \mathbb{R}^{1\times D}$ is concatenated with $\boldsymbol{T}_{out}$ for an input $\boldsymbol{X}_{0}\in \mathbb{R}^{(N_{L_1}+1)\times D}$ of this stage. Notably, the [{\tt CLASS}] part would only be processed by the attention branch.
\begin{align}
    &\boldsymbol{X}_{0}=\left[\boldsymbol{T}_{out}; \boldsymbol{C}_0\right]
\end{align}
Denoting the ASF-C with function $f_{ac}\left(\cdot\right)$, the process of the $j$-th encoders is:
\begin{align}
    &\boldsymbol{X}_{j}=f_{ac}\left(\boldsymbol{X}_{j-1}\right), \quad \boldsymbol{X}_{j}\in \mathbb{R}^{(N_{L_1}+1)\times D}
\end{align}

The [{\tt CLASS}] token yielded by the last ASF-C encoders will be fed into a fully-connected layer for category prediction:
\begin{align}
    &\boldsymbol{Y}=Linear\left(\boldsymbol{C}_{L_2}\right), \quad \boldsymbol{Y}\in \mathbb{R}^{Categories}
\end{align}

Since ASF-R/C encoders share most parts, we present them together in Section~\ref{sec:enc}. 

\subsection{An ASF-R/C Encoder}
\label{sec:enc}
The ASF-R \& ASF-C encoders are same in \textit{Split Parallelism}, \textit{Adaptive Fusion} and MLP parts, and differs in the attention part (T2T or vanilla attention).

\textbf{Split Parallelism} equally split a tensor of tokens $\boldsymbol{T}\in\mathbb{R}^{N\times D}$ for the ASF-R (or $\boldsymbol{X}$ for the ASF-C) into two parts $\boldsymbol{T}^{(a)}$, $\boldsymbol{T}^{(b)} \in \mathbb{R}^{N \times \frac{D}{2}}$, along the channel axis. Then, the sub-tensor $\boldsymbol{T}^{(a)}$/$\boldsymbol{T}^{(b)}$ is separately fed into convolutional/attention branch for local/global modeling. Notably, $\boldsymbol{T}^{(a)}$ are pre/post-processed with \textit{seq2image} or \textit{image2seq} function \cite{yuan2021tokens} to re-arrange tokens into spatial or sequential form. The process is shown below:
\begin{align}
    &\boldsymbol{\hat{T}}^{(a)}=img2seq\left(f_{convb}\left(seq2img\left(\boldsymbol{T}^{(a)}\right)\right)\right)\\
    &\boldsymbol{\hat{T}}^{(b)}=f_{atteb}\left(\boldsymbol{T}^{(b)}\right)
\end{align}
where $f_{atteb}(\cdot)$ and $f_{convb}(\cdot)$ respectively denote attention and convolution paths, and $\boldsymbol{\hat{T}}^{(a)}$,  $\boldsymbol{\hat{T}}^{(b)}\in \mathbb{R}^{N\times D'}$. Hereby, $D'$=64  in the ASF-R (or $D'$=$\frac{D}{2}$ in ASF-C). Notably, we carefully craft an efficient convolutional branch named \textit{Half-Residual Mobile Convolutional Branch} and present it in Section~\ref{sec:conv}. 

\textbf{Adaptive Fusion} performs weighted sum on tensors processed by the two paths with adaptive scalars $\alpha$ and $\beta$. Hereby, $\alpha$ and $\beta$ are calculated according to visual features from the two paths by Eq.~(\ref{eq:alpha})$\sim$(\ref{eq:beta}). 
\begin{align}
    &\boldsymbol{S}=  \boldsymbol{\hat{T}}^{(a)} + \boldsymbol{\hat{T}}^{(b)} \\
    & \alpha= Sigmoid\left(f_{w}\left(\boldsymbol{S}\right)\right)\label{eq:alpha}\\
    & \beta = 1-\alpha\label{eq:beta}\\
    &\boldsymbol{\hat{T}}= \alpha \cdot \boldsymbol{\hat{T}}^{(a)} + \beta \cdot \boldsymbol{\hat{T}}^{(b)} + \boldsymbol{S}
\end{align}
where the $f_w\left(\cdot\right)$ denotes the function for generating weighting scalars. Notably, we generate the $\alpha$ \& $\beta$ in a \textit{Sigmoid} way. Though this way is theoretically equivalent to a \textit{Softmax} function, it is practically simple in implementation. We describe details and compare different fusion strategies in Section~\ref{sec:adaptive}.

\textbf{Attentions \& MLP} are mostly inherited from the general vision transformer regime, with minor modifications on attention settings. Specifically, the ASF-R/C separately adopt the T2T attention and vanilla attentions. Compared with the vanilla, the T2T attention replaces the multi-head scheme with a single-head one and fixes channels of ``query'', ``key'', ``value'' to $D'=64$, concerning computational efficiency. Since the T2T attention reshapes tokens, the shortcut and Conv $1\times1$ are removed in the ASF-R compared with the ASF-C ({\color{red} red} line in Fig.~(\ref{fig:framework})). Output $\boldsymbol{\widetilde{T}}$/$\boldsymbol{\widetilde{X}}$ of the ASF-R/C encoders is generated as in Eq.~(\ref{eq:outr})$\sim$(\ref{eq:outc}), where $f_{mlp}\left(\cdot\right)$ denotes the MLP with two \textit{fc} layers and a GeLU activation:
\begin{align}
    &\boldsymbol{\widetilde{T}}= f_{mlp}\left(\boldsymbol{\hat{T}}\right) + \boldsymbol{\hat{T}}\label{eq:outr}\\
    &\boldsymbol{\widetilde{X}}= f_{mlp}\left(\boldsymbol{\mathring{X}}\right) + \boldsymbol{\mathring{X}}, \qquad \boldsymbol{\mathring{X}}= Conv\left(\boldsymbol{\hat{X}}\right) + \boldsymbol{X}\label{eq:outc}
\end{align}

\subsection{Half-Residual Mobile Convolutional Branch}
\label{sec:conv}
\begin{figure}
\centering
\includegraphics[height=4.5cm]{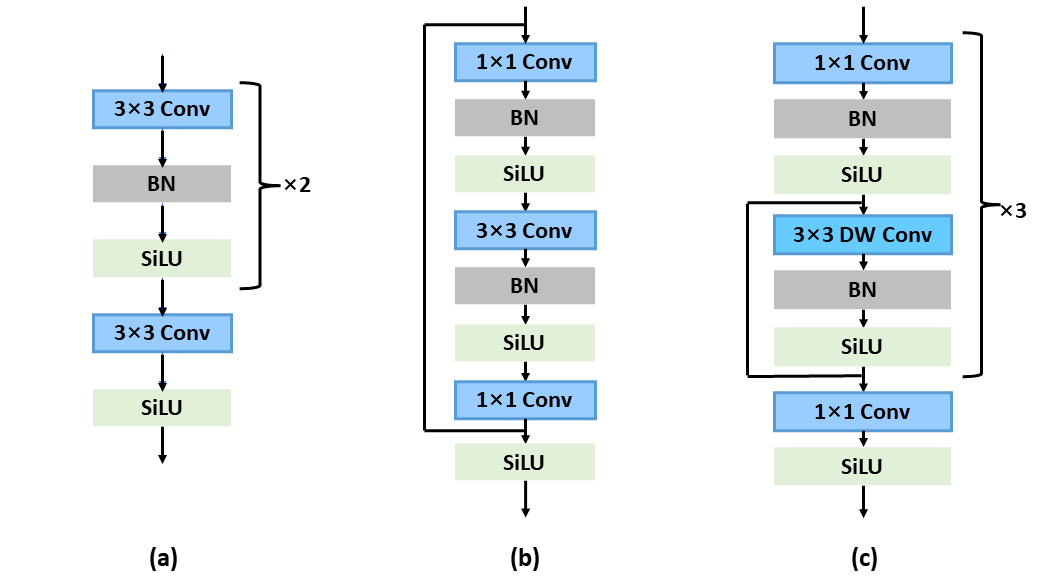}
\caption{Designs of convolutional branch: (a) PCM from ViTAE transformer\cite{Xu2021ViTAEVT}; (b) Residual bottleneck; (c) our \textbf{HMCB}}
\label{fig:convbranch}
\end{figure}
We study existing CNN branches for hybrid models and craft a new \textit{Half-Residual Mobile Convolutional Branch} (HMCB). The HMCB is more complementary to the attention way than its counterparts while consuming fewer computations. We begin with PCM, the recently proposed CNN-branch in ViTAE hybrid \cite{Xu2021ViTAEVT}.

\textbf{Parallel Convolution Module} (PCM) is shown in Figure~(\ref{fig:convbranch}a), it contains stacked convolutions with $3\times3$ kernels. We re-implement it by replacing its internal \textit{group conv} with conventional \textit{conv}. 
This re-implementation is combined with our Split Parallelism and surpasses the original ViTAE hybrid (Accuracy from 82.0\% $\rightarrow$ 82.2\%), at the expense of extra parameters.

\textbf{Residual Bottleneck} is widely used in various vision tasks (Fig.~(\ref{fig:convbranch}b)). Thus, we want to adopt it as the convolutional branch in the hybrid models. We only implement one block of residual bottleneck here to maintain similar params/MACs as the original PCM.

\textbf{Half-Residual Mobile Convolutional Branch} (HMCB) is modified based on PCM, with the help of efficient \textit{conv} approximations (Fig.~(\ref{fig:convbranch}c)). Inspired by MobileNet \cite{howard2017mobilenets} and MobileNetV2 \cite{Sandler2018MobileNetV2IR}, 
we first factorize each conventional $3\times3$ \textit{conv} into one $3\times3$ \textit{depth-wise conv} followed by one $1\times1$ \textit{conv} and then we add another $1\times1$ \textit{conv} before the first \textit{depth-wise conv}.

These approximations remarkably reduce computations. Even if we replicate the half-residual block three times, the HMCB still contains similar Params / MACs to single Residual bottleneck. Specifically, we implant the shortcut at a different position with the conventional residual bottleneck to be compatible with the repetition and promote the training across channels. We compare the three designs in terms of accuracy, Params, and MACs in Table~\ref{table:convbranch} and observe that our HMCB performs the best under all metrics.

\subsection{Adaptive Fusion and Counterparts}
\label{sec:adaptive}
In this part, we present Adaptive Fusion and two simple counterparts. We begin with a \textit{simple fusion} with fixed weights, then introduce a fusion strategy \textit{context-agnostic} weights, and finally, give the \textit{Adaptive Fusion} with contextually relevant weights.

\begin{figure}
\centering
\includegraphics[height=4.5cm]{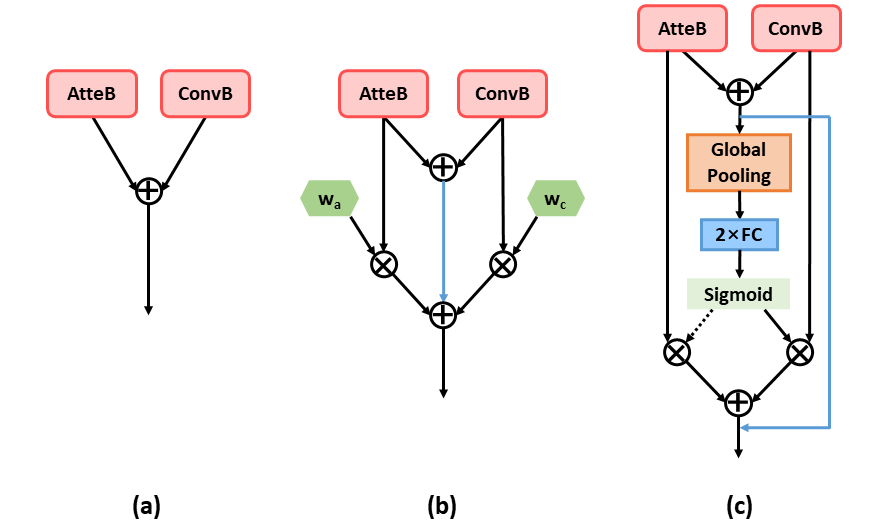}
\caption{Fusion strategies: (a) Simple Fusion; (b) Context-Agnostic Fusion; (c) Adaptive Fusion. The shortcut ({\color{blue}blue}) serves for reducing gradient vanishment}
\label{fig:fusion}
\end{figure}

\textbf{Simple Fusion} directly averages outputs from the two branches with equal importance as in Eq.~(\ref{eq:sf}) and Fig.~(\ref{fig:fusion}a). The fusion itself is parameter-free and effective. Thus, it is preferred in a pilot parallel hybrid, ViTAE \cite{Xu2021ViTAEVT}. 
\begin{align}
    & \alpha= \beta = 0.5 \label{eq:sf}
\end{align}

\textbf{Context-Agnostic Fusion} explicitly learns $\alpha$ \& $\beta$ on par with training process (Fig.~(\ref{fig:fusion}b)). To avoid a phenomenon that the gradient vanishment deactivates a particular branch when $W_\alpha$ or $W_\beta$ falls into extremely tiny values, we add an extra skip connection ({\color{blue}blue} line) to enforce gradients to be propagated to both ways.
\begin{align}
    & \alpha= W_\alpha,\quad \beta = W_\beta
\end{align}

\textbf{Adaptive Fusion} calculates $\alpha$ and $\beta$ according to visual contexts from both branches, which is inspired by SKNet~\cite{Li2019SelectiveKN}. Its process is shown in Eq.~(\ref{eq:af1})\&(\ref{eq:af2}) and Figure~(\ref{fig:fusion}c).  
\begin{equation}
    \alpha= Sigmoid\left(Linear^2\left(Pool\left(\boldsymbol{\hat{T}}^{(a)} + \boldsymbol{\hat{T}}^{(b)}\right)\right)\right)
    \label{eq:af1}
\end{equation}
\begin{equation}
    \beta = 1-\alpha
    \label{eq:af2}
\end{equation}
Specifically, we expand the function $f_w$ in Eq. (\ref{eq:alpha}) to be two fully-connected layers (\textit{Linear}), with \textit{BatchNorm} and \textit{GeLU} activations in between. To stabilize the training procedure, we add the extra skip connection from the same concern as the prior fusion method.

We comprehensively compare various fusion strategies in Table~\ref{table:fusion} and observe a significant improvement with our Adaptive Fusion strategy.

\subsection{Pyramid Structure}
\label{pyramid}

\begin{figure*}
\centering
\includegraphics[height=4.5cm]{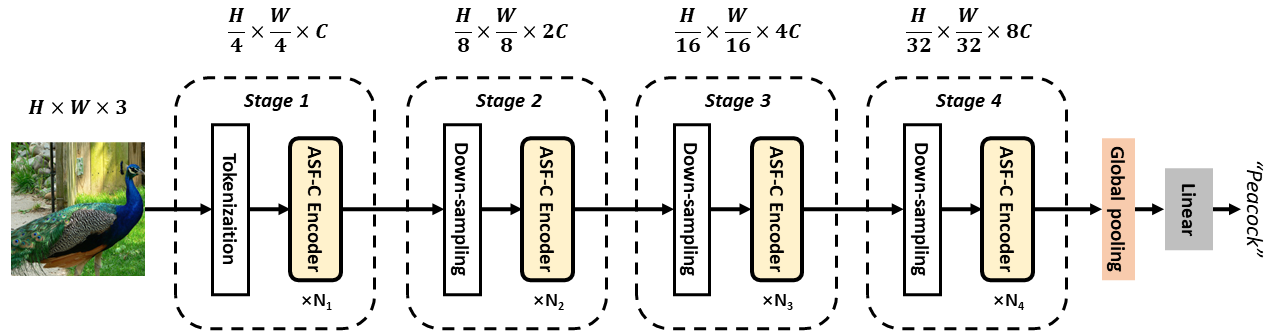}
\caption{The architecture of ASF-former$_p$. It adopts a 4-stage pyramid structure and is built based on ASF-C encoders. Without [{\tt CLASS}] token, the final output feature will be entirely fed into the classification layer after global average pooling.}
\label{fig:pyramid}
\end{figure*}

ASF-former is built beyond the structure of T2T-ViT~\cite{yuan2021tokens}, which has a reduction stage for down-sampling and a computation stage for feature encoding under the same feature dimension. However, downstream tasks such as object detection, instance segmentation, and semantic segmentation usually prefer multi-level features from the backbone model to capture objects at different scales. This requirement makes ASF-former not quite suitable for transferring to downstream tasks. To deal with this problem, we follow the pyramid structure of Swin Transformer~\cite{liu2021Swin} and further design a new model with our ASF-C encoder, denoted as ASF-former$_p$.

As Figure~(\ref{fig:pyramid}) shows, ASF-former$_p$ adopts a 4-stage pyramid structure with down-sampling layers before each stage. The original images are tokenized and down-sampled by the ratio of 4 in the first stage, while the other down-sampling layers use a ratio of 2. ASF-former$_p$ has $N_1$,$N_2$,$N_3$,$N_4$ ASF-C encoders in each stage respectively and the feature channel dimension $C$ in Stage 1 will be doubled in each following stages. Following Swin~\cite{liu2021Swin}, we use the simple convolutions for image tokenization and down-sampling instead of T2T operation; therefore, we could remove the ASF-R from the ASF-former$_p$. Besides, instead of adding a [{\tt CLASS}] token for classification, we perform global average pooling on the output feature of the last stage and directly feed it into a fully-connected layer for category prediction.

We will compare ASF-former and its pyramid version, ASF-former$_p$, in Table~\ref{table:sota}. Besides, in Section~\ref{detection} and \ref{segmentation}, we will adopt ASF-former$_p$ for all downstream tasks and further utilize its stage-wise nature.

\section{Experiments}
We evaluate the ASF-former on standard benchmarks, including ImageNet-1K, CIFAR-10/100, with metrics like Top-1/5 accuracy, model Params, and inferencing MACs. Experimental results validate the efficacy and efficiency of the ASF-former.
\subsection{Datasets}
We conduct ablation and transferability experiments on ImageNet-1K and CIFAR-10/100 downstream datasets.

\textbf{ImageNet-1K} \cite{imagenet_cvpr09} defines 1000 categories with 1.35 million images captured in daily life. On average, each category contains around 1.3k samples. These images are split into training/validation sets with a ratio of 26:1.

\textbf{CIFAR-10/100} \cite{krizhevsky2009learning} respectively contains 10/100 categories on 60k images with a fixed $32\times32$ resolution. In the CIFAR-10, each category includes 6k samples, with 5k/1k samples as training/testing. Whereas, in the CIFAR-100, there are 600 samples per category, with 500/100 for training/testing.

\subsection{Experimental settings}
In this part, we briefly introduce hyperparameters of our ASF-former variants and the training \& inference recipes.

\setlength{\tabcolsep}{8pt}
\begin{table*}
\begin{center}
\caption{Hyperparameters of ASF-former-S \& ASF-former-B .}
\label{table:variant}
\begin{tabular}{l|ccc|ccc|rr}
\hline\noalign{\smallskip}
\multirow{3}*{Model} &\multicolumn{3}{c}{Reduction stage} &\multicolumn{3}{c}{Computation stage} &\multicolumn{2}{c}{Model Size}\\
 \cline{2-9}
 &Depth &Token &MLP &Depth &Token &MLP &Params &MACs\\
 &$L_1$ &dim $D'$ &dim &$L_2$ &dim $D$ &dim &(M) &(G)\\
\noalign{\smallskip}
\hline
\noalign{\smallskip}
\textbf{ASF-former-S} &2 &64 &64 &14 &384 &1152 &\textbf{19.3} &\textbf{5.5}\\
\textbf{ASF-former-B} &2 &64 &64 &24 &512 &1536 &\textbf{56.7} &\textbf{12.9}\\
\hline
\end{tabular}
\end{center}
\end{table*}
\setlength{\tabcolsep}{1.4pt}

\textbf{ASF-former variants}. By customizing hyperparameters, such as the number of encoders (i.e., $L_1$ and $L_2$) and dimensions of tokens in different layers, we can flexibly control the complexity of ASF-former at different computation scales. To fairly compare the ASF-former with its counterpart of similar computational costs, we propose a \textit{small} and \textit{big} model, respectively denoted as the ASF-former-S and ASF-former-B in Table~\ref{table:variant}. Besides, we set the same $k$, $o$, $p$ as the original T2T-ViT model (Eq.~(\ref{patch_number})). As for ASF-former$_p$, we also design two variants correspondingly, named ASF-former$_p$-S and ASF-former$_p$-B. Their details are shown in Table~\ref{table:variant_p}.

\setlength{\tabcolsep}{8pt}
\begin{table}
\begin{center}
\caption{Hyperparameters of ASF-former$_p$-S \& ASF-former$_p$-B .}
\label{table:variant_p}
\begin{tabular}{l|cc|rr}
\hline\noalign{\smallskip}
\multirow{2}*{Model} &Depth &Dim &Params &MACs\\
 &$N_1$,$N_2$,$N_3$,$N_4$ &$C$ &(M) &(G)\\
\noalign{\smallskip}
\hline
\noalign{\smallskip}
\textbf{ASF-former$_p$-S} &3,4,12,5 &64 &\textbf{21.3} &\textbf{5.4}\\
\textbf{ASF-former$_p$-B} &3,4,18,5 &96 &\textbf{58.9} &\textbf{12.1}\\
\hline
\end{tabular}
\end{center}
\end{table}
\setlength{\tabcolsep}{1.4pt}

\textbf{Training \& Inference}. We fix the training/inference recipe as \cite{yuan2021tokens} for a fair comparison. In the training phase, images are randomly cropped into size $224\times224$ before going through the network. We also adopt data-augmentations, such as MixUp \cite{zhang2018mixup}, CutMix \cite{yun2019cutmix}, Rand-Augment \cite{Cubuk2020RandaugmentPA}, Random-Erasing \cite{zhong2020random} to reduce over-fitting. The Exponential Moving Average (EMA) strategy is further used for training stability. We train 310 epochs using AdamW optimization, with a batch size of 512. The learning rate is initialized with 5e-4 and decreases with the cosine learning schedule. 
In the inference phase, images are first resized to let the short side be 256 and then center-cropped into $224 \times 224$ before being fed into the network.

\subsection{Ablation study}
\label{sec:abla}
In this part, we study the effectiveness of our proposed convolutional branch HMCB, Split Parallelism, Adaptive Fusion, etc. We test them on top of the \textit{small} ASF-former-S for quick verification. 

\textbf{HMCB} \textit{vs} \textbf{Convolutional Candidates}. We plug the PCM, Residual Bottleneck, and HMCB into the ASF-former. To exclude the influence of the fusion strategy, we employ ``Simple Fusion'' in all three ASF-formers. The comparison is shown in Table~\ref{table:convbranch}.
\setlength{\tabcolsep}{4pt}
\begin{table}[th]
\begin{center}
\caption{Comparison of different convolutional branchs on ImageNet-1K Val.}
\label{table:convbranch}
\begin{tabular}{llcccc}
\hline\noalign{\smallskip}
Conv Branch &Regime &Params (M)&MACs (G) &Top-1 (\%)\\
\noalign{\smallskip}
\hline
\noalign{\smallskip}
PCM &ViTAE \cite{Xu2021ViTAEVT}& 23.6 & 5.6 & 82.0 \\
PCM &ASF-former& 32.1 & 9.3 & 82.2 \\
Residual Bottleneck&ASF-former& \textbf{18.3} & \textbf{5.4} & 81.7\\
Our HMCB&ASF-former& 18.8 & 5.5 & \textbf{82.5} \\
\hline
\end{tabular}
\end{center}
\end{table}
\setlength{\tabcolsep}{1.4pt}

We observe that the HMCB achieves the best accuracy (82.5\%) among all candidates while consuming comparable or fewer computations (Params / MACs) than the Residual Bottleneck or PCM. This validates that the HMCB is more complementary to global attention than the rest at a low computational cost. Moreover, plugging PCM into ASF-former (with simple fusion) performs slightly better than in the original ViTAE, verifying the effectiveness of the Split Parallelism mechanism. We fix the convolutional branch to be an HMCB in the following experiments.

\textbf{Split Parallelism} \textit{vs} \textbf{Single Path}. We further compare the Split Parallelism with Single Path methods. We remove the channel split for the Single Path method and feed the entire input into an Attention-Only or HMCB-Only path. Hereby,  we still adopt ``Simple Fusion'' (Fig.~(\ref{fig:fusion}a)) in this ablation . Notably, the HMCB-only replaces the [{\tt CLASS}] token with an average pooled vector to predict final categories.
\setlength{\tabcolsep}{4pt}
\begin{table}[th]
\begin{center}
\caption{Comparison of Split Parallelism and Sing Path on ImageNet-1K Val.}
\label{table:branch}
\begin{tabular}{lcccc}
\hline\noalign{\smallskip}
Branch &Params (M)&MACs (G)&Top-1 (\%)\\
\noalign{\smallskip}
\hline
\noalign{\smallskip}
Attention-only & 21.5 & 6.1 & 81.7\\
HMCB-only & 22.7 & \textbf{5.2} & 72.4\\
\textbf{Attention + HMCB}& \textbf{18.8} & 5.5 & \textbf{82.5} \\
\hline
\end{tabular}
\end{center}
\end{table}
\setlength{\tabcolsep}{1.4pt}

The results are shown in Table~\ref{table:branch}. Our Split Parallelism achieves 82.5\% accuracy, which remarkably outperforms single-path settings (81.7\% for Atten-only and 72.4\% for Conv-only). Thanks to the Split strategy, our parallelism achieves comparable or fewer Parameters \& MACs than single path methods. This also indicates that the HMCB and attention branches are complementary; meanwhile, our Split Parallelism could capture and integrate the information from both branches very well.

\textbf{Adaptive Fusion} \textit{vs} \textbf{Counterparts}. We implement fusion strategies in Section~\ref{sec:adaptive}, including ``Simple Fusion'', ``Context-Agnostic Fusion'' and ``Adaptive Fusion'', on top of the ASF-former. All fusion variants intake outputs from the attention branch and HMCB. We present their comparison in Table~\ref{table:fusion}.
\setlength{\tabcolsep}{4pt}
\begin{table}[th]
\begin{center}
\caption{Comparison of different fusion method on ImageNet validation set.}
\label{table:fusion}
\begin{tabular}{lcccc}
\hline\noalign{\smallskip}
Fusion Method &Params &MACs &Top-1\\
\noalign{\smallskip}
\hline
\noalign{\smallskip}
Simple Fusion & 18.8 & 5.5 & 82.5 \\
Context-Agnostic Fusion & 18.8 & 5.5 & 82.2 \\
\textbf{Adaptive Fusion} & 19.3 & 5.5 & \textbf{82.7} \\
\hline
\end{tabular}
\end{center}
\end{table}
\setlength{\tabcolsep}{1.4pt}

We find that our Adaptive Fusion achieves 82.7\% accuracy, which is superior to all the other counterparts under similar parameters and MACs. This indicates the effectiveness of adapting the weights according to visual contents and verifies the different branch preferences of different images. Notably, Context-Agnostic Fusion performs worse than Simple Fusion, showing that the coarsely learning context-agnostic weights would even degrade both branches' capability and training effect.

\textbf{Effectiveness of Shortcut}. We validate the influence of the shortcut ({\color{blue}blue} line in Figure~(\ref{fig:fusion}c)) by removing it from Adaptive Fusion. The comparison is shown in Table~\ref{table:shortcut}.
\setlength{\tabcolsep}{4pt}
\begin{table}[th]
\begin{center}
\caption{Effectiveness of shortcut on ImageNet Val.}
\label{table:shortcut}
\begin{tabular}{lccl}
\hline\noalign{\smallskip}
Fusion Method &Params (M)&MACs (G) &Top-1 (\%)\\
\noalign{\smallskip}
\hline
\noalign{\smallskip}
ASF-Former-S & 19.3 & 5.5 & 82.7 \\
~~~~$-$ shortcut & 19.3 & 5.5 & 82.0 ($\downarrow$ 0.7) \\
\hline
\end{tabular}
\end{center}
\end{table}
\setlength{\tabcolsep}{1.4pt}

When discarding the skip connection, we can see that the final accuracy degrades by a large margin (0.7\%) and is even much worse than Simple Fusion in Table~\ref{table:fusion}. This demonstrates the necessity of skip connection when fusing the outputs of two branches and verifies its ability to help the model's training by promoting gradient propagation.

\textbf{Effectiveness of Pyramid Structure}. We train and validate the pyramid version of ASF-former-S \&ASF-former-B, named ASF-former$_p$-S \& ASF-former$_p$-B, under the same recipe. The results are present in Table~\ref{table:sota}.

The underlined ASF-former$_p$-S achieves an excellent accuracy of 83.0\%, which outperforms ASF-former-S by 0.3\% with comparable parameters and MACs. This strongly proves the effectiveness of a multi-level pyramid structure for better feature encoding when designing hybrid transformer models. ASF-former$_p$-B achieves comparable results with ASF-former-B while consuming fewer MACs. This further indicates the computation-efficient property of our pyramid model.

\subsection{Comparison with the state-of-the-art}

\setlength{\tabcolsep}{4pt}
\begin{table}
\begin{center}
\caption{Comparison with different methods on ImageNet validation set.}
\label{table:sota}
\begin{tabular}{l|l|crr|cc}
\hline\noalign{\smallskip}
\multirow{2}*{Type} &\multirow{2}*{Model} &Image &Params &MACs &\multicolumn{2}{c}{ImageNet}\\
 & &Size &(M) &(G) &Top-1 &Top-5\\
\noalign{\smallskip}
\midrule[1pt]
\noalign{\smallskip}
\multirow{3}*{\rotatebox{90}{\makecell[c]{\textit{CNN}\\(Small)}}} &ResNet-50 \cite{He2016DeepRL} & 224 & 25.6 & 7.6 & 76.7 & 93.3\\
 &RegNetY-4G \cite{Radosavovic2020DesigningND} & 224 & 21.0 & 4.0 & 80.0 & -\\
 &ConvNeXt-T \cite{liu2022convnet} & 224 & 29.0 & 4.5 & 82.1 & -\\
\midrule
\multirow{6}*{\rotatebox{90}{\makecell[c]{\textit{Transformer}\\(Small)}}} &PVT-S \cite{wang2021pyramid} & 224 & 24.5 & 7.6 & 79.8 & -\\
 &DeiT-S \cite{Touvron2021TrainingDI} & 224 & 22.0 & 4.6 & 79.9 & 95.0\\
 &Swin-T \cite{liu2021Swin} & 224 & 28.0 & 4.5 & 81.2 & 95.5\\
 &TNT-S \cite{tnt} & 224 & 23.8 & 5.2 & 81.5 & 95.7\\
 &T2T-ViT-14 \cite{yuan2021tokens} & 224 & 21.5 & 5.2 & 81.5 & 95.7\\
 &T2T-ViT$_t$-14 \cite{yuan2021tokens} & 224 & 21.5 & 6.1 & 81.7 & -\\
\midrule
\multirow{12}*{\rotatebox{90}{\makecell[c]{\textit{Hybrid}\\(Small)}}}  &ConT-M \cite{Yan2021ConTNetWN} & 224 & 19.2 & 3.1 & 80.2 & -\\
 &ConT-B \cite{Yan2021ConTNetWN} & 224 & 39.6 & 6.4 & 81.8 & -\\
 &Conformer-Ti \cite{Peng2021ConformerLF} & 224 & 23.5 & 5.2 & 81.3 & -\\
 &Swin-ACmix-T \cite{Pan2021OnTI} & 224 & 30.0 & 4.6 & 81.9 & -\\
 &CeiT-S \cite{Yuan2021IncorporatingCD} & 224 & 24.2 & 4.5 & 82.0 & 95.9\\
 &ViTAE-S \cite{Xu2021ViTAEVT} & 224 & 23.6 & 5.6 & 82.0 & 95.9\\
 &DAT-T \cite{Xia2022CVPR_DAT} & 224 & 29.0 & 4.6 & 82.0 & -\\
 &Focal-T \cite{Yang2021FocalSF} & 224 & 29.1 & 4.9 & 82.2 & -\\
 &CvT-13 \cite{Wu2021CvTIC} & 224 & 20.0 & 4.5 & 81.6 & -\\
 &CvT-21 \cite{Wu2021CvTIC} & 224 & 32.0 & 7.1 & 82.5 & -\\
 &\textbf{ASF-former-S} & 224 & 19.3 & 5.5 & \textbf{82.7} & \textbf{96.1}\\
 &\textbf{\uline{ASF-former$_p$-S}} & 224 & 21.3 & 5.4 & \textbf{\uline{83.0}} & \textbf{\uline{96.3}}\\
\midrule[1.0pt]
 \multirow{4}*{\rotatebox{90}{\makecell[c]{\textit{CNN}\\(Big)}}} 
 &ResNet-101 \cite{He2016DeepRL} & 224 & 44.5 & 15.2 & 78.3 & 94.1\\
 &ResNet-152 \cite{He2016DeepRL} & 224 & 60.2 & 22.6 & 78.9 & 94.4\\
 &RegNetY-16G \cite{Radosavovic2020DesigningND} & 224 & 84.0 & 16.0 & 82.9 & -\\
 &ConvNeXt-B \cite{liu2022convnet} & 224 & 89.0 & 15.4 & 83.8 & -\\
\midrule
\multirow{8}*{\rotatebox{90}{\makecell[c]{\textit{Transformer}\\(Big)}}} &ViT-B/16 \cite{dosovitskiy2021an} & 384 & 86.5 & 55.4 & 77.9 & -\\
 &ViT-L/16 \cite{dosovitskiy2021an} & 384 & 304.3 & 65.8 & 76.5 & -\\
 &PVT-L \cite{wang2021pyramid} & 224 & 61.4 & 19.6 & 81.7 & -\\
 &DeiT-B \cite{Touvron2021TrainingDI} & 224 & 86.6 & 34.6 & 81.8 & 95.6\\
 &T2T-ViT-24 \cite{yuan2021tokens} & 224 & 64.1 & 14.1 & 82.3 & -\\
 &T2T-ViT$_t$-24 \cite{yuan2021tokens} & 224 & 64.1 & 15.0 & 82.6 & -\\
 &TNT-B \cite{tnt} & 224 & 65.6 & 14.1 & 82.9 & 96.3\\
 &Swin-B \cite{liu2021Swin} & 224 & 88.0 & 15.4 & 83.5 & 96.5\\
\midrule
\multirow{5}*{\rotatebox{90}{\makecell[c]{\textit{Hybrid}\\(Big)}}}  
 &Conformer-S \cite{Peng2021ConformerLF} & 224 & 37.7 & 10.6 & 83.4 & -\\
 &Swin-ACmix-S \cite{Pan2021OnTI} & 224 & 51.0 & 9.0 & 83.5 & -\\
 &Focal-B \cite{Yang2021FocalSF} & 224 & 89.8 & 16.0 & 83.8 & -\\
 &\textbf{ASF-former-B} & 224 & 56.7 & 12.9 & \textbf{83.9} & \textbf{96.6}\\
 &\textbf{\uline{ASF-former$_p$-B}} & 224 & 58.9 & 12.1 & \textbf{\uline{83.9}} & \textbf{\uline{96.5}}\\
\hline
\end{tabular}
\end{center}
\end{table}
\setlength{\tabcolsep}{1.4pt}

\begin{figure*}
\centering
\includegraphics[height=5.5cm]{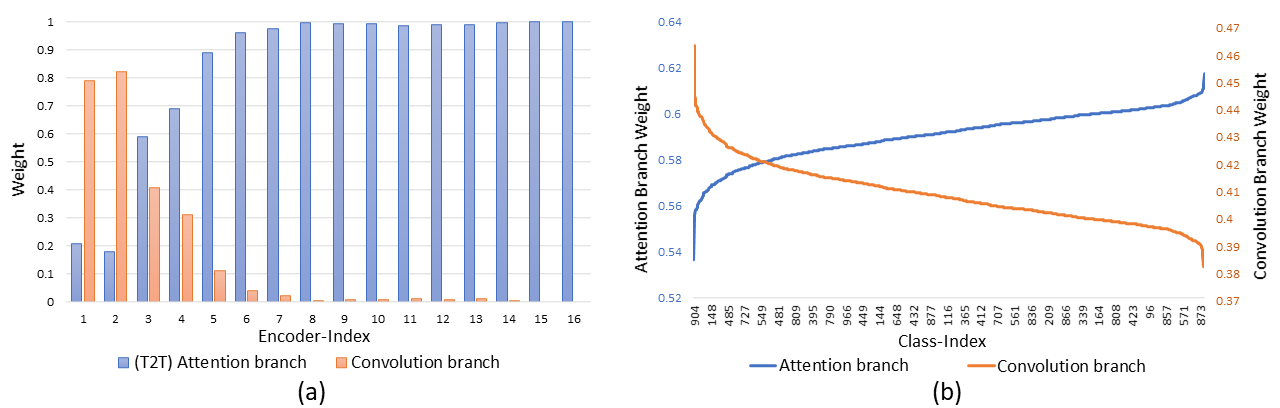}
\caption{The distribution of weights for HMCB and Attention branch. (a) Weights to the depth of encoder. (b) Weights to categories. (Blue/Orange denotes weights for the Attention/HMCB, this figure is best viewed in color)}
\label{fig:vis}
\end{figure*}

We further compare the ASF-former-S and ASF-former-B with SOTAs of pure CNN, transformer, and CNN-transformer hybrid. We separately present models into \textit{Small} and \textit{Big} parts, considering their computation scales (Params/MACs). We only analyzed the results of ASF-former below for brevity. However, it has to be mentioned that ASF-former$_p$ performs comparably with or even better than ASF-former and further expands the advantages of our design, as shown in Table~\ref{table:sota}.

\textbf{Compared with CNN SOTAs}, our ASF-former outperforms the strong ConvNeXt regime in terms of accuracy with fewer parameters and comparable computations. For example, the {ASF-former-S} is better than the ConvNeXt-T ({82.7}\% \textit{vs} 82.1\%) in accuracy, with much fewer parameters ({19.3}M \textit{vs} 29.0M) and slightly more computations ({5.5}G \textit{vs} 4.5G). And, the {ASF-former-B} surpasses the ConvNeXt-B ({83.9}\% \textit{vs} 83.8\%) with much less parameters ({56.7}M \textit{vs} 89.0M) and MACs ({12.9}G \textit{vs} 15.4G). 

\textbf{Compared with pure transformer SOTAs}, the ASF-former performs better than the Swin Transformer and T2T-ViT regimes in terms of accuracy, Params, with comparable MACs. Specifically, the ASF-former-S achieves higher accuracy than the Swin-T and  T2T-ViT$_t$-14 (82.7\% \textit{vs} 81.2\% \textit{vs} 81.7\%), with fewer parameters (19.3M \textit{vs} 28.0M \textit{vs} 21.5M) and comparable MACs (5.5M \textit{vs} 4.5M \textit{vs} 6.1M). Besides, the ASF-former-B outperforms the Swin-B and T2T-ViT$_t$-24 under all metrics: Accuracy (83.9\% \textit{vs} 83.5\% \textit{vs} 82.6\%), Params (56.7M \textit{vs} 88M \textit{vs} 64.1M), MACs (12.9G \textit{vs} 15.4G \textit{vs} 15.0G).

\textbf{Compared with the CNN-transformer hybrid SOTAs}, the ASF-former outperforms those cascade hybrids, such as CvT and Focal Transformer, in terms of accuracy and Params, becoming the first parallel hybrid to beat cascade counterparts. Specifically, at a similar MAC scale, the ASF-former-S shows a better accuracy (82.7\% \textit{vs} 82.5\% \textit{vs} 82.2\%) and fewer params (19.3M \textit{vs} 32.0M \textit{vs} 29.1M) than CvT-21 and Focal-T. Meanwhile, the ASF-former-B is better than Focal-B under all metrics: Accuracy (83.9\% \textit{vs} 83.8\%), Params (56.7M \textit{vs} 89.8M), MACs (12.9G \textit{vs} 16.0G).

Among hybrid transformers, ViTAE, Swin-ACmix, Conformer, and our ASF-former all adopt parallel structure, while the ASF-former demonstrates the best accuracy at a similar computation scale. For example, the ASF-former-S and ASF-former-B separately surpass those best available, i.e., ViTAE-S and Swin-ACmix-S, by an accuracy margin of 82.7\%-82.0\%=0.7\% and 83.9\%-83.5\%=0.4\%. This indicates that our split parallelism with HMCB, cooperating with the adaptive fusion, can efficiently enforce the model to be lightweight and effectively boost performance via integrating parallel features.

\setlength{\tabcolsep}{4pt}
\begin{table}
\begin{center}
\caption{Comparison with different methods on ImageNet-1k validation set after ImageNet-22k pre-training.}
\label{table:22k}
\begin{tabular}{lcccc}
\hline\noalign{\smallskip}
\multirow{2}*{Model} &Image &Params &MACs &ImageNet\\
 &Size &(M) &(G) &Top-1\\
\noalign{\smallskip}
\hline
\noalign{\smallskip}
R-101x3~\cite{Kolesnikov2020BigT}& 384 &388 &204.6 &84.4\\
ViT-B/16~\cite{dosovitskiy2021an} &384 &86 &55.4 &84.0\\
ViT-L/16~\cite{dosovitskiy2021an} &384 &307 &190.7 &\textbf{85.2}\\
Swin-B~\cite{liu2021Swin}&224 &88 &15.4 &\textbf{85.2}\\
\midrule
\textbf{ASF-Former-S} (ours) &224 &\textbf{56.7} &\textbf{12.9} &\textbf{85.2}\\
\hline
\end{tabular}
\end{center}
\end{table}
\setlength{\tabcolsep}{1.4pt}

\textbf{Compared with SOTAs under ImageNet-22k pre-training}, we also pre-train the model on the large-scale ImageNet-22k dataset and then fine-tune it on ImageNet-1k to further explore the upper bound of ASF-former. The classification results are shown in Table~\ref{table:22k}. Our ASF-former achieves the same Top-1 accuracy (85.2\%) as the strong SOTA of Swin-B but requires much less parameters (56.7M \textit{vs} 88.0M) and MACs (12.9G \textit{vs} 15.4G). These indicate the great capacity of the ASF-former as well as its superior computation efficiency.

\subsection{Distribution of Weights}

We plot the distribution of weights for HMCB and Attention branch with respect to the depth of encoder and categories in Fig.~\ref{fig:vis}(a) and (b). For simplicity, we calculate distributions using ASF-former-S.

Figure~(\ref{fig:vis}a) shows the trend of weights changing with the depth of the encoder. Specifically, the ASF-former-S contains 16 encoders. For each encoder, we calculate the mean weight of the HMCB/Attention way on ImageNet-1K Val. We observe that the domination of HMCB in early encoders gradually changes to the attention when depth becomes deeper. This finding is consistent with prior works \cite{Pan2021OnTI} that shallow layers focus on locality, whereas deep layers prefer globality, which will inspire future model designing.

Figure~(\ref{fig:vis}b) shows distribution weights on 1000 categories. We pick the third encoders as they are the most balanced for the HMCB/Attention (3rd depth in Fig.~(\ref{fig:vis}a)). We sort categories according to the descending (or increment) of HMCB (Attention) weight. We observe that categories prefer locality/globality differently, which is also affected by the depth of the encoder.
Specifically, small objects like “\textit{Ruler}” and “\textit{Can opener}” prefer the locality. Categories like “\textit{Honeycomb}” also produce large weights for the convolution branch for its requirement of capturing the local texture \& edge information. Sceneries or large objects like “\textit{Triumphal arch}”, "\textit{Kimono}" and "\textit{Leonberg}" prefer global information.

\subsection{Transferability to downstream datasets}

\setlength{\tabcolsep}{4pt}
\begin{table}
\begin{center}
\caption{Transferability to CIFAR-10/100.}
\label{table:generalization}
\begin{tabular}{lcccc}
\hline\noalign{\smallskip}
Model &Params &CIFAR-10 &CIFAR-100\\
\noalign{\smallskip}
\hline
\noalign{\smallskip}
ViT-B/16 \cite{dosovitskiy2021an} & 86.5 & 98.1 & 87.1\\
ViT-L/16 \cite{dosovitskiy2021an} & 304.3 & 97.9 & 86.4\\
T2T-ViT-14 \cite{yuan2021tokens} &21.5 &97.5 &88.4 \\
TNT-S$_{\uparrow384}$ \cite{tnt} &23.8 &98.7 &90.1\\
ViTAE-S$_{\uparrow384}$ \cite{Xu2021ViTAEVT} &23.6 &98.8 &90.8 \\
CeiT-S$_{\uparrow384}$ \cite{Yuan2021IncorporatingCD}  &24.2 &\textbf{99.1} &90.8\\
DeiT-B$_{\uparrow384}$ \cite{Touvron2021TrainingDI} & 86.6 & \textbf{99.1} & 90.8\\

\midrule
\textbf{ASF-former-S} &19.3 &98.7 &90.4 \\
\textbf{ASF-former-B} &56.7 &98.8 &\textbf{91.0} \\
\hline
\end{tabular}
\end{center}
\end{table}
\setlength{\tabcolsep}{1.4pt}

To investigate the transferability of our ASF-former, we further fine-tune the proposed models on CIFAR-10 and CIFAR-100 datasets. The initial learning rate is 0.025 for CIFAR-10 and 0.05 for CIFAR-100.

The validation results are shown in Table~\ref{table:generalization}. Our ASF-former achieves comparable results on CIFAR-10 and the state-of-the-art results on CIFAR-100 under $224 \times 224$ resolution, showing its superior transferability.

\subsection{Transferability to object detection \& instance segmentation}
\label{detection}

\setlength{\tabcolsep}{4pt}
\begin{table}
\begin{center}
\caption{Transferability to object detection and instance segmentation task.}
\label{table:detection}
\begin{tabular}{llcccc}
\hline\noalign{\smallskip}
\multirow{2}*{Backbone} &\multirow{2}*{Method} &Lr &Params &Box &Mask\\
& &Schd &(M) &mAP &mAP\\
\noalign{\smallskip}
\hline
\noalign{\smallskip}
ResNet-50~\cite{He2016DeepRL}&Mask RCNN&1x&44&38.2&34.7\\
PVT-S~\cite{wang2021pyramid}&Mask RCNN &1x &44 &40.4 &37.8\\
ResT-Base~\cite{Zhang2021ResTAE}&Mask RCNN &1x &50 &41.6 &38.7\\
Twins-PCSVT-S~\cite{Chu2021TwinsRT}&Mask RCNN &1x &44 &42.9 &40.0\\
Twins-PVT-S~\cite{Chu2021TwinsRT}&Mask RCNN &1x &44 &43.4 &40.3\\
RegionViT-S+~\cite{Chen2021RegionViTRA}&Mask RCNN &1x &51 &43.5 &40.4\\
Conformer-S~\cite{Peng2021ConformerLF}&Mask RCNN &1x &58 &43.6 &39.7\\
Swin-T~\cite{liu2021Swin}&Mask RCNN&1x&48&43.7&39.8 \\
DAT-T~\cite{Xia2022CVPR_DAT}&Mask RCNN&1x&48 &44.4 &40.4\\
ViTAE-S~\cite{Xu2021ViTAEVT}&Mask RCNN&1x&\textbf{37}& 44.6&40.2\\
Focal-T~\cite{Yang2021FocalSF}&Mask RCNN&1x&49 &44.8 &41.0\\
\textbf{ASF-Former$_p$-S} (ours)&Mask RCNN &1x &38 &\textbf{45.9} &\textbf{41.2}\\
\midrule
ResNet-50~\cite{He2016DeepRL}&Cascade RCNN &1x &82 &41.2 &35.9\\
Swin-T~\cite{liu2021Swin}&Cascade RCNN &1x &86 &48.1 &41.7 \\   
ViTAE-S~\cite{Xu2021ViTAEVT}&Cascade RCNN &1x &\textbf{75} &48.9 &42.0\\
DAT-T~\cite{Xia2022CVPR_DAT}&Cascade RCNN &1x &86 &49.1 &42.5\\
\textbf{ASF-Former$_p$-S} (ours)&Cascade RCNN &1x &79 &\textbf{49.7} &\textbf{42.8}\\
\hline
\end{tabular}
\end{center}
\end{table}
\setlength{\tabcolsep}{1.4pt}

As described in Section~\ref{pyramid}, ASF-former$_p$ is adjusted from the ASF-former, to make the transformer more compatible with the detection \& segmentation tasks. We test the performance of our ASF-former$_p$ on standard object detection and instance segmentation benchmarks: MS-COCO 2017~\cite{Lin2014Mscoco} datasets. Specifically, the COCO contains 118k training images and 5k validation images. During experiments, we adopt Mask RCNN and Cascade RCNN as the detection frameworks and follow the same experimental settings as Swin~\cite{liu2021Swin} for a fair comparison. Specifically, multi-scale training and AdamW optimizer are used. The initial learning rate is set to be 0.0001 with a weight decay of 0.05. We use the batch size of 16 and 1x schedule (12 epochs).

The experimental results are shown in Table~\ref{table:detection}. We compare our ASF-former$_p$ with different kinds of vision backbones, including the most widely-used pure CNN, transformer, and the recent strong hybrid models. Among them, ASF-former$_p$ achieves the best results on both object detection (Box mAP) and instance segmentation (Mask mAP) tasks under two widely-used detection frameworks with comparable or fewer parameters. It indicates the superior capacity of ASF-former$_p$ as a backbone for downstream tasks.

\subsection{Transferability to semantic segmentation}
\label{segmentation}

\setlength{\tabcolsep}{4pt}
\begin{table}
\begin{center}
\caption{Transferability to semantic segmentation task.}
\label{table:segmentation}
\begin{tabular}{llcccc}
\hline\noalign{\smallskip}
\multirow{2}*{Backbone} &\multirow{2}*{Method} &Lr &Params &\multirow{2}*{mIoU} &mIoU\\
& &Schd &(M) & &(MS)\\
\noalign{\smallskip}
\hline
\noalign{\smallskip}
ResNet-50~\cite{He2016DeepRL}&UPerNet &160K &67 &42.1 &42.9\\
Swin-T~\cite{liu2021Swin}&UPerNet &160K &60 &44.5 &45.8\\
Swin-ACmix-T~\cite{Pan2021OnTI}&UPerNet &160K &60 &45.3 &-\\
DAT-T~\cite{Xia2022CVPR_DAT}&UPerNet &160K &60 &45.5 &46.4\\
ViTAE-S~\cite{Xu2021ViTAEVT}&UPerNet &160K &\textbf{49} &45.4 &\textbf{47.8}\\
DW-T~\cite{Ren2022BeyondFD}&UPerNet &160K &61 &45.7 &46.9\\
Focal-T~\cite{Yang2021FocalSF}&UPerNet &160K &62 &45.8 &47.0\\
ConvNeXt-T~\cite{liu2022convnet}&UPerNet &160K &60 &46.0 &46.7\\
Twins-SVT-S~\cite{Chu2021TwinsRT}&UPerNet &160K &54 &46.2 &47.1\\
Twins-PCPVT-S~\cite{Chu2021TwinsRT}&UPerNet &160K &55 &46.2 &47.5\\
\textbf{ASF-Former$_p$-S} (ours)&UPerNet &160K &51 &\textbf{46.7} &47.6\\
\hline
\end{tabular}
\end{center}
\end{table}
\setlength{\tabcolsep}{1.4pt}

We further validate our ASF-former$_p$ as the backbone model for semantic segmentation tasks using the standard ADE20K~\cite{Zhou2019ade20k} dataset. It contains more than 20k images in the training set and 2k images in the validation set. Following Swin, we take the widely-used UPerNet as the framework and use the AdamW optimizer for training. The initial learning rate is set to be 0.00006 with a weight decay of 0.01, and the batch size is 16. We follow the 160K lr schedule (160k iterations) for a fair comparison.

The results are shown in Table~\ref{table:segmentation} which include the standard and multi-scale (MS) testing results. We can see that ASF-former$_p$ achieves a mIoU of 46.7 in standard testing, which outperforms the other backbone models by a large margin with comparable or fewer parameters. Under a multi-scale augmentation setting, ASF-former$_p$ gets comparable results to the SOTA methods. These show that our method works well as the backbone in segmentation tasks and proves its excellent transferability to downstream vision tasks.

\section{Conclusion}
In this paper, we propose a novel hybrid transformer called ASF-former. It adopts Split Parallelism, which splits channels into half for two-path inputs. It introduces the novel HMCB that complements the attention and the Adaptive Fusion for feature merging. We experimentally verified that the three mechanisms could a good balance of efficacy and efficiency and achieve SOTA results. We also validate that the role of local/global information varies with respect to visual categories and network depth. Besides, this hybrid design also achieves promising results on downstream tasks, including object detection and segmentation.

\normalem
\bibliographystyle{IEEEtran}
\bibliography{tmm}

\begin{thebibliography}{10}
\providecommand{\url}[1]{#1}
\csname url@samestyle\endcsname
\providecommand{\newblock}{\relax}
\providecommand{\bibinfo}[2]{#2}
\providecommand{\BIBentrySTDinterwordspacing}{\spaceskip=0pt\relax}
\providecommand{\BIBentryALTinterwordstretchfactor}{4}
\providecommand{\BIBentryALTinterwordspacing}{\spaceskip=\fontdimen2\font plus
\BIBentryALTinterwordstretchfactor\fontdimen3\font minus
  \fontdimen4\font\relax}
\providecommand{\BIBforeignlanguage}[2]{{%
\expandafter\ifx\csname l@#1\endcsname\relax
\typeout{** WARNING: IEEEtran.bst: No hyphenation pattern has been}%
\typeout{** loaded for the language `#1'. Using the pattern for}%
\typeout{** the default language instead.}%
\else
\language=\csname l@#1\endcsname
\fi
#2}}
\providecommand{\BIBdecl}{\relax}
\BIBdecl

\bibitem{yuan2021tokens}
L.~Yuan, Y.~Chen, T.~Wang, W.~Yu, Y.~Shi, Z.~Jiang, F.~E.~H. Tay, J.~Feng, and
  S.~Yan, ``Tokens-to-token vit: Training vision transformers from scratch on
  imagenet,'' in \emph{2021 {IEEE/CVF} International Conference on Computer
  Vision, {ICCV} 2021, Montreal, QC, Canada, October 10-17, 2021}, 2021, pp.
  538--547.

\bibitem{dosovitskiy2021an}
A.~Dosovitskiy, L.~Beyer, A.~Kolesnikov, D.~Weissenborn, X.~Zhai,
  T.~Unterthiner, M.~Dehghani, M.~Minderer, G.~Heigold, S.~Gelly, J.~Uszkoreit,
  and N.~Houlsby, ``An image is worth 16x16 words: Transformers for image
  recognition at scale,'' in \emph{9th International Conference on Learning
  Representations, {ICLR} 2021, Virtual Event, Austria, May 3-7, 2021}, 2021.

\bibitem{Chen2021VisformerTV}
Z.~Chen, L.~Xie, J.~Niu, X.~Liu, L.~Wei, and Q.~Tian, ``Visformer: The
  vision-friendly transformer,'' in \emph{2021 {IEEE/CVF} International
  Conference on Computer Vision, {ICCV} 2021, Montreal, QC, Canada, October
  10-17, 2021}, 2021, pp. 569--578.

\bibitem{Hassani2021EscapingTB}
A.~Hassani, S.~Walton, N.~Shah, A.~Abuduweili, J.~Li, and H.~Shi, ``Escaping
  the big data paradigm with compact transformers,'' \emph{CoRR}, vol.
  abs/2104.05704, 2021.

\bibitem{Mehta2021MobileViTLG}
S.~Mehta and M.~Rastegari, ``Mobilevit: Light-weight, general-purpose, and
  mobile-friendly vision transformer,'' \emph{CoRR}, vol. abs/2110.02178, 2021.

\bibitem{Wu2021CvTIC}
H.~Wu, B.~Xiao, N.~Codella, M.~Liu, X.~Dai, L.~Yuan, and L.~Zhang, ``Cvt:
  Introducing convolutions to vision transformers,'' in \emph{2021 {IEEE/CVF}
  International Conference on Computer Vision, {ICCV} 2021, Montreal, QC,
  Canada, October 10-17, 2021}, 2021, pp. 22--31.

\bibitem{Yan2021ConTNetWN}
H.~Yan, Z.~Li, W.~Li, C.~Wang, M.~Wu, and C.~Zhang, ``Contnet: Why not use
  convolution and transformer at the same time?'' \emph{CoRR}, vol.
  abs/2104.13497, 2021.

\bibitem{Yang2021FocalSF}
J.~Yang, C.~Li, P.~Zhang, X.~Dai, B.~Xiao, L.~Yuan, and J.~Gao, ``Focal
  self-attention for local-global interactions in vision transformers,''
  \emph{CoRR}, vol. abs/2107.00641, 2021.

\bibitem{Yuan2021IncorporatingCD}
K.~Yuan, S.~Guo, Z.~Liu, A.~Zhou, F.~Yu, and W.~Wu, ``Incorporating convolution
  designs into visual transformers,'' in \emph{2021 {IEEE/CVF} International
  Conference on Computer Vision, {ICCV} 2021, Montreal, QC, Canada, October
  10-17, 2021}, 2021, pp. 559--568.

\bibitem{He2021PruningSI}
H.~He, J.~Liu, Z.~Pan, J.~Cai, J.~Zhang, D.~Tao, and B.~Zhuang, ``Pruning
  self-attentions into convolutional layers in single path,'' \emph{CoRR}, vol.
  abs/2111.11802, 2021.

\bibitem{Chen2021MobileFormerBM}
Y.~Chen, X.~Dai, D.~Chen, M.~Liu, X.~Dong, L.~Yuan, and Z.~Liu,
  ``Mobile-former: Bridging mobilenet and transformer,'' \emph{CoRR}, vol.
  abs/2108.05895, 2021.

\bibitem{Pan2021OnTI}
X.~Pan, C.~Ge, R.~Lu, S.~Song, G.~Chen, Z.~Huang, and G.~Huang, ``On the
  integration of self-attention and convolution,'' \emph{CoRR}, vol.
  abs/2111.14556, 2021.

\bibitem{Peng2021ConformerLF}
Z.~Peng, W.~Huang, S.~Gu, L.~Xie, Y.~Wang, J.~Jiao, and Q.~Ye, ``Conformer:
  Local features coupling global representations for visual recognition,'' in
  \emph{2021 {IEEE/CVF} International Conference on Computer Vision, {ICCV}
  2021, Montreal, QC, Canada, October 10-17, 2021}, 2021, pp. 357--366.

\bibitem{Xu2021ViTAEVT}
Y.~Xu, Q.~Zhang, J.~Zhang, and D.~Tao, ``Vitae: Vision transformer advanced by
  exploring intrinsic inductive bias,'' in \emph{Advances in Neural Information
  Processing Systems 34: Annual Conference on Neural Information Processing
  Systems 2021, NeurIPS 2021, December 6-14, 2021, virtual}, 2021, pp.
  28\,522--28\,535.

\bibitem{Krizhevsky2012ImageNetCW}
A.~Krizhevsky, I.~Sutskever, and G.~E. Hinton, ``Imagenet classification with
  deep convolutional neural networks,'' in \emph{Advances in Neural Information
  Processing Systems 25: 26th Annual Conference on Neural Information
  Processing Systems 2012. Proceedings of a meeting held December 3-6, 2012,
  Lake Tahoe, Nevada, United States}, 2012, pp. 1106--1114.

\bibitem{Simonyan2015VeryDC}
K.~Simonyan and A.~Zisserman, ``Very deep convolutional networks for
  large-scale image recognition,'' in \emph{3rd International Conference on
  Learning Representations, {ICLR} 2015, San Diego, CA, USA, May 7-9, 2015,
  Conference Track Proceedings}, 2015.

\bibitem{Szegedy2015GoingDW}
C.~Szegedy, W.~Liu, Y.~Jia, P.~Sermanet, S.~E. Reed, D.~Anguelov, D.~Erhan,
  V.~Vanhoucke, and A.~Rabinovich, ``Going deeper with convolutions,'' in
  \emph{{IEEE} Conference on Computer Vision and Pattern Recognition, {CVPR}
  2015, Boston, MA, USA, June 7-12, 2015}, 2015, pp. 1--9.

\bibitem{Szegedy2016RethinkingTI}
C.~Szegedy, V.~Vanhoucke, S.~Ioffe, J.~Shlens, and Z.~Wojna, ``Rethinking the
  inception architecture for computer vision,'' in \emph{2016 {IEEE} Conference
  on Computer Vision and Pattern Recognition, {CVPR} 2016, Las Vegas, NV, USA,
  June 27-30, 2016}, 2016, pp. 2818--2826.

\bibitem{He2016DeepRL}
K.~He, X.~Zhang, S.~Ren, and J.~Sun, ``Deep residual learning for image
  recognition,'' in \emph{2016 {IEEE} Conference on Computer Vision and Pattern
  Recognition, {CVPR} 2016, Las Vegas, NV, USA, June 27-30, 2016}, 2016, pp.
  770--778.

\bibitem{He2016IdentityMI}
------, ``Identity mappings in deep residual networks,'' in \emph{Computer
  Vision - {ECCV} 2016 - 14th European Conference, Amsterdam, The Netherlands,
  October 11-14, 2016, Proceedings, Part {IV}}, vol. 9908, 2016, pp. 630--645.

\bibitem{Huang2017DenselyCC}
G.~Huang, Z.~Liu, L.~van~der Maaten, and K.~Q. Weinberger, ``Densely connected
  convolutional networks,'' in \emph{2017 {IEEE} Conference on Computer Vision
  and Pattern Recognition, {CVPR} 2017, Honolulu, HI, USA, July 21-26, 2017},
  2017, pp. 2261--2269.

\bibitem{Girshick2015FastR}
R.~B. Girshick, ``Fast {R-CNN},'' in \emph{2015 {IEEE} International Conference
  on Computer Vision, {ICCV} 2015, Santiago, Chile, December 7-13, 2015}, 2015,
  pp. 1440--1448.

\bibitem{Ren2015FasterRT}
S.~Ren, K.~He, R.~B. Girshick, and J.~Sun, ``Faster {R-CNN:} towards real-time
  object detection with region proposal networks,'' in \emph{Advances in Neural
  Information Processing Systems 28: Annual Conference on Neural Information
  Processing Systems 2015, December 7-12, 2015, Montreal, Quebec, Canada},
  2015, pp. 91--99.

\bibitem{Redmon2016YouOL}
J.~Redmon, S.~K. Divvala, R.~B. Girshick, and A.~Farhadi, ``You only look once:
  Unified, real-time object detection,'' in \emph{2016 {IEEE} Conference on
  Computer Vision and Pattern Recognition, {CVPR} 2016, Las Vegas, NV, USA,
  June 27-30, 2016}, 2016, pp. 779--788.

\bibitem{Liu2016SSDSS}
W.~Liu, D.~Anguelov, D.~Erhan, C.~Szegedy, S.~E. Reed, C.~Fu, and A.~C. Berg,
  ``{SSD:} single shot multibox detector,'' in \emph{Computer Vision - {ECCV}
  2016 - 14th European Conference, Amsterdam, The Netherlands, October 11-14,
  2016, Proceedings, Part {I}}, vol. 9905, 2016, pp. 21--37.

\bibitem{Lin2017FocalLF}
T.~Lin, P.~Goyal, R.~B. Girshick, K.~He, and P.~Doll{\'{a}}r, ``Focal loss for
  dense object detection,'' in \emph{{IEEE} International Conference on
  Computer Vision, {ICCV} 2017, Venice, Italy, October 22-29, 2017}, 2017, pp.
  2999--3007.

\bibitem{Lin2017FeaturePN}
T.~Lin, P.~Doll{\'{a}}r, R.~B. Girshick, K.~He, B.~Hariharan, and S.~J.
  Belongie, ``Feature pyramid networks for object detection,'' in \emph{2017
  {IEEE} Conference on Computer Vision and Pattern Recognition, {CVPR} 2017,
  Honolulu, HI, USA, July 21-26, 2017}, 2017, pp. 936--944.

\bibitem{Cai2018CascadeRD}
Z.~Cai and N.~Vasconcelos, ``Cascade {R-CNN:} delving into high quality object
  detection,'' in \emph{2018 {IEEE} Conference on Computer Vision and Pattern
  Recognition, {CVPR} 2018, Salt Lake City, UT, USA, June 18-22, 2018}, 2018,
  pp. 6154--6162.

\bibitem{He2017MaskR}
K.~He, G.~Gkioxari, P.~Doll{\'{a}}r, and R.~B. Girshick, ``Mask {R-CNN},'' in
  \emph{{IEEE} International Conference on Computer Vision, {ICCV} 2017,
  Venice, Italy, October 22-29, 2017}, 2017, pp. 2980--2988.

\bibitem{Chen2017RethinkingAC}
L.~Chen, G.~Papandreou, F.~Schroff, and H.~Adam, ``Rethinking atrous
  convolution for semantic image segmentation,'' \emph{CoRR}, vol.
  abs/1706.05587, 2017.

\bibitem{Chen2018DeepLabSI}
L.~Chen, G.~Papandreou, I.~Kokkinos, K.~Murphy, and A.~L. Yuille, ``Deeplab:
  Semantic image segmentation with deep convolutional nets, atrous convolution,
  and fully connected crfs,'' \emph{{IEEE} Trans. Pattern Anal. Mach. Intell.},
  vol.~40, no.~4, pp. 834--848, 2018.

\bibitem{Feichtenhofer2016ConvolutionalTN}
C.~Feichtenhofer, A.~Pinz, and A.~Zisserman, ``Convolutional two-stream network
  fusion for video action recognition,'' in \emph{2016 {IEEE} Conference on
  Computer Vision and Pattern Recognition, {CVPR} 2016, Las Vegas, NV, USA,
  June 27-30, 2016}, 2016, pp. 1933--1941.

\bibitem{Qiu2017LearningSR}
Z.~Qiu, T.~Yao, and T.~Mei, ``Learning spatio-temporal representation with
  pseudo-3d residual networks,'' in \emph{{IEEE} International Conference on
  Computer Vision, {ICCV} 2017, Venice, Italy, October 22-29, 2017}, 2017, pp.
  5534--5542.

\bibitem{Carreira2017QuoVA}
J.~Carreira and A.~Zisserman, ``Quo vadis, action recognition? {A} new model
  and the kinetics dataset,'' in \emph{2017 {IEEE} Conference on Computer
  Vision and Pattern Recognition, {CVPR} 2017, Honolulu, HI, USA, July 21-26,
  2017}, 2017, pp. 4724--4733.

\bibitem{Zhou2018TemporalRR}
B.~Zhou, A.~Andonian, A.~Oliva, and A.~Torralba, ``Temporal relational
  reasoning in videos,'' in \emph{Computer Vision - {ECCV} 2018 - 15th European
  Conference, Munich, Germany, September 8-14, 2018, Proceedings, Part {I}},
  vol. 11205, 2018, pp. 831--846.

\bibitem{Wang2018NonlocalNN}
X.~Wang, R.~B. Girshick, A.~Gupta, and K.~He, ``Non-local neural networks,'' in
  \emph{2018 {IEEE} Conference on Computer Vision and Pattern Recognition,
  {CVPR} 2018, Salt Lake City, UT, USA, June 18-22, 2018}, 2018, pp.
  7794--7803.

\bibitem{Feichtenhofer2019SlowFastNF}
C.~Feichtenhofer, H.~Fan, J.~Malik, and K.~He, ``Slowfast networks for video
  recognition,'' in \emph{2019 {IEEE/CVF} International Conference on Computer
  Vision, {ICCV} 2019, Seoul, Korea (South), October 27 - November 2, 2019},
  2019, pp. 6201--6210.

\bibitem{Lin2019TSMTS}
J.~Lin, C.~Gan, and S.~Han, ``{TSM:} temporal shift module for efficient video
  understanding,'' in \emph{2019 {IEEE/CVF} International Conference on
  Computer Vision, {ICCV} 2019, Seoul, Korea (South), October 27 - November 2,
  2019}, 2019, pp. 7082--7092.

\bibitem{howard2017mobilenets}
A.~G. Howard, M.~Zhu, B.~Chen, D.~Kalenichenko, W.~Wang, T.~Weyand,
  M.~Andreetto, and H.~Adam, ``Mobilenets: Efficient convolutional neural
  networks for mobile vision applications,'' \emph{CoRR}, vol. abs/1704.04861,
  2017.

\bibitem{Sandler2018MobileNetV2IR}
M.~Sandler, A.~G. Howard, M.~Zhu, A.~Zhmoginov, and L.~Chen, ``Mobilenetv2:
  Inverted residuals and linear bottlenecks,'' in \emph{2018 {IEEE} Conference
  on Computer Vision and Pattern Recognition, {CVPR} 2018, Salt Lake City, UT,
  USA, June 18-22, 2018}, 2018, pp. 4510--4520.

\bibitem{Zhang2018ShuffleNetAE}
X.~Zhang, X.~Zhou, M.~Lin, and J.~Sun, ``Shufflenet: An extremely efficient
  convolutional neural network for mobile devices,'' in \emph{2018 {IEEE}
  Conference on Computer Vision and Pattern Recognition, {CVPR} 2018, Salt Lake
  City, UT, USA, June 18-22, 2018}, 2018, pp. 6848--6856.

\bibitem{Tan2019EfficientNetRM}
M.~Tan and Q.~V. Le, ``Efficientnet: Rethinking model scaling for convolutional
  neural networks,'' in \emph{Proceedings of the 36th International Conference
  on Machine Learning, {ICML} 2019, 9-15 June 2019, Long Beach, California,
  {USA}}, vol.~97, 2019, pp. 6105--6114.

\bibitem{Hu2018SqueezeandExcitationN}
J.~Hu, L.~Shen, and G.~Sun, ``Squeeze-and-excitation networks,'' in \emph{2018
  {IEEE} Conference on Computer Vision and Pattern Recognition, {CVPR} 2018,
  Salt Lake City, UT, USA, June 18-22, 2018}, 2018, pp. 7132--7141.

\bibitem{Hu2018GatherExciteEF}
J.~Hu, L.~Shen, S.~Albanie, G.~Sun, and A.~Vedaldi, ``Gather-excite: Exploiting
  feature context in convolutional neural networks,'' in \emph{Advances in
  Neural Information Processing Systems 31: Annual Conference on Neural
  Information Processing Systems 2018, NeurIPS 2018, December 3-8, 2018,
  Montr{\'{e}}al, Canada}, 2018, pp. 9423--9433.

\bibitem{hao2022group}
Y.~Hao, H.~Zhang, C.-W. Ngo, and X.~He, ``Group contextualization for video
  recognition,'' in \emph{Proceedings of the IEEE/CVF Conference on Computer
  Vision and Pattern Recognition}, 2022, pp. 928--938.

\bibitem{Li2019SelectiveKN}
X.~Li, W.~Wang, X.~Hu, and J.~Yang, ``Selective kernel networks,'' in
  \emph{{IEEE} Conference on Computer Vision and Pattern Recognition, {CVPR}
  2019, Long Beach, CA, USA, June 16-20, 2019}, 2019, pp. 510--519.

\bibitem{Zhang2020ResNeStSN}
H.~Zhang, C.~Wu, Z.~Zhang, Y.~Zhu, Z.~Zhang, H.~Lin, Y.~Sun, T.~He, J.~Mueller,
  R.~Manmatha, M.~Li, and A.~J. Smola, ``Resnest: Split-attention networks,''
  \emph{CoRR}, vol. abs/2004.08955, 2020.

\bibitem{alfasly2022effective}
S.~Alfasly, C.~K. Chui, Q.~Jiang, J.~Lu, and C.~Xu, ``An effective video
  transformer with synchronized spatiotemporal and spatial self-attention for
  action recognition,'' \emph{IEEE Transactions on Neural Networks and Learning
  Systems}, 2022.

\bibitem{EAPT}
X.~Lin, S.~Sun, W.~Huang, B.~Sheng, P.~Li, and D.~D. Feng, ``Eapt: Efficient
  attention pyramid transformer for image processing,'' \emph{IEEE Transactions
  on Multimedia}, pp. 1--1, 2021.

\bibitem{YDTR}
W.~Tang, F.~He, and Y.~Liu, ``Ydtr: Infrared and visible image fusion via
  y-shape dynamic transformer,'' \emph{IEEE Transactions on Multimedia}, pp.
  1--16, 2022.

\bibitem{Li2021ExploitingMP}
M.~Li, J.~Liu, C.~Zheng, X.~Huang, and Z.~Zhang, ``Exploiting multi-view
  part-wise correlation via an efficient transformer for vehicle
  re-identification,'' \emph{IEEE Transactions on Multimedia}, pp. 1--1, 2021.

\bibitem{Zhao2022SpatialET}
J.~Zhao, H.~Wang, Y.~Zhou, R.~Yao, S.~Chen, and A.~El~Saddik, ``Spatial-channel
  enhanced transformer for visible-infrared person re-identification,''
  \emph{IEEE Transactions on Multimedia}, pp. 1--1, 2022.

\bibitem{zhou2022moving}
Y.~Zhou, Y.~Wang, and L.-P. Chau, ``Moving towards centers: Re-ranking with
  attention and memory for re-identification,'' \emph{IEEE Transactions on
  Multimedia}, 2022.

\bibitem{FT-TDR}
J.~Wang, S.~Chen, Z.~Wu, and Y.-G. Jiang, ``Ft-tdr: Frequency-guided
  transformer and top-down refinement network for blind face inpainting,''
  \emph{IEEE Transactions on Multimedia}, pp. 1--1, 2022.

\bibitem{Li2022ExploitingTC}
W.~Li, H.~Liu, R.~Ding, M.~Liu, P.~Wang, and W.~Yang, ``Exploiting temporal
  contexts with strided transformer for 3d human pose estimation,'' \emph{IEEE
  Transactions on Multimedia}, pp. 1--1, 2022.

\bibitem{Touvron2021TrainingDI}
H.~Touvron, M.~Cord, M.~Douze, F.~Massa, A.~Sablayrolles, and H.~J{\'{e}}gou,
  ``Training data-efficient image transformers {\&} distillation through
  attention,'' in \emph{Proceedings of the 38th International Conference on
  Machine Learning, {ICML} 2021, 18-24 July 2021, Virtual Event}, vol. 139,
  2021, pp. 10\,347--10\,357.

\bibitem{tnt}
K.~Han, A.~Xiao, E.~Wu, J.~Guo, C.~Xu, and Y.~Wang, ``Transformer in
  transformer,'' in \emph{Advances in Neural Information Processing Systems 34:
  Annual Conference on Neural Information Processing Systems 2021, NeurIPS
  2021, December 6-14, 2021, virtual}, 2021, pp. 15\,908--15\,919.

\bibitem{wang2021pyramid}
W.~Wang, E.~Xie, X.~Li, D.~Fan, K.~Song, D.~Liang, T.~Lu, P.~Luo, and L.~Shao,
  ``Pyramid vision transformer: {A} versatile backbone for dense prediction
  without convolutions,'' in \emph{2021 {IEEE/CVF} International Conference on
  Computer Vision, {ICCV} 2021, Montreal, QC, Canada, October 10-17, 2021},
  2021, pp. 548--558.

\bibitem{liu2021Swin}
Z.~Liu, Y.~Lin, Y.~Cao, H.~Hu, Y.~Wei, Z.~Zhang, S.~Lin, and B.~Guo, ``Swin
  transformer: Hierarchical vision transformer using shifted windows,'' in
  \emph{2021 {IEEE/CVF} International Conference on Computer Vision, {ICCV}
  2021, Montreal, QC, Canada, October 10-17, 2021}, 2021, pp. 9992--10\,002.

\bibitem{imagenet_cvpr09}
J.~Deng, W.~Dong, R.~Socher, L.~Li, K.~Li, and L.~Fei{-}Fei, ``Imagenet: {A}
  large-scale hierarchical image database,'' in \emph{2009 {IEEE} Computer
  Society Conference on Computer Vision and Pattern Recognition {(CVPR} 2009),
  20-25 June 2009, Miami, Florida, {USA}}, 2009, pp. 248--255.

\bibitem{krizhevsky2009learning}
A.~Krizhevsky, G.~Hinton \emph{et~al.}, ``Learning multiple layers of features
  from tiny images,'' 2009.

\bibitem{zhang2018mixup}
H.~Zhang, M.~Ciss{\'{e}}, Y.~N. Dauphin, and D.~Lopez{-}Paz, ``mixup: Beyond
  empirical risk minimization,'' in \emph{6th International Conference on
  Learning Representations, {ICLR} 2018, Vancouver, BC, Canada, April 30 - May
  3, 2018, Conference Track Proceedings}, 2018.

\bibitem{yun2019cutmix}
S.~Yun, D.~Han, S.~Chun, S.~J. Oh, Y.~Yoo, and J.~Choe, ``Cutmix:
  Regularization strategy to train strong classifiers with localizable
  features,'' in \emph{2019 {IEEE/CVF} International Conference on Computer
  Vision, {ICCV} 2019, Seoul, Korea (South), October 27 - November 2, 2019},
  2019, pp. 6022--6031.

\bibitem{Cubuk2020RandaugmentPA}
E.~D. Cubuk, B.~Zoph, J.~Shlens, and Q.~Le, ``Randaugment: Practical automated
  data augmentation with a reduced search space,'' in \emph{Advances in Neural
  Information Processing Systems 33: Annual Conference on Neural Information
  Processing Systems 2020, NeurIPS 2020, December 6-12, 2020, virtual}, 2020.

\bibitem{zhong2020random}
Z.~Zhong, L.~Zheng, G.~Kang, S.~Li, and Y.~Yang, ``Random erasing data
  augmentation,'' in \emph{The Thirty-Fourth {AAAI} Conference on Artificial
  Intelligence, {AAAI} 2020, New York, NY, USA, February 7-12, 2020}, 2020, pp.
  13\,001--13\,008.

\bibitem{Radosavovic2020DesigningND}
I.~Radosavovic, R.~P. Kosaraju, R.~B. Girshick, K.~He, and P.~Doll{\'{a}}r,
  ``Designing network design spaces,'' in \emph{2020 {IEEE/CVF} Conference on
  Computer Vision and Pattern Recognition, {CVPR} 2020, Seattle, WA, USA, June
  13-19, 2020}, 2020, pp. 10\,425--10\,433.

\bibitem{liu2022convnet}
Z.~Liu, H.~Mao, C.-Y. Wu, C.~Feichtenhofer, T.~Darrell, and S.~Xie, ``A convnet
  for the 2020s,'' \emph{Proceedings of the IEEE/CVF Conference on Computer
  Vision and Pattern Recognition (CVPR)}, 2022.

\bibitem{Xia2022CVPR_DAT}
Z.~Xia, X.~Pan, S.~Song, L.~E. Li, and G.~Huang, ``Vision transformer with
  deformable attention,'' in \emph{Proceedings of the IEEE/CVF Conference on
  Computer Vision and Pattern Recognition (CVPR)}, June 2022, pp. 4794--4803.

\bibitem{Kolesnikov2020BigT}
A.~Kolesnikov, L.~Beyer, X.~Zhai, J.~Puigcerver, J.~Yung, S.~Gelly, and
  N.~Houlsby, ``Big transfer (bit): General visual representation learning,''
  in \emph{Computer Vision - {ECCV} 2020 - 16th European Conference, Glasgow,
  UK, August 23-28, 2020, Proceedings, Part {V}}, vol. 12350, 2020, pp.
  491--507.

\bibitem{Zhang2021ResTAE}
Q.~Zhang and Y.~Yang, ``Rest: An efficient transformer for visual
  recognition,'' in \emph{Advances in Neural Information Processing Systems 34:
  Annual Conference on Neural Information Processing Systems 2021, NeurIPS
  2021, December 6-14, 2021, virtual}, 2021, pp. 15\,475--15\,485.

\bibitem{Chu2021TwinsRT}
X.~Chu, Z.~Tian, Y.~Wang, B.~Zhang, H.~Ren, X.~Wei, H.~Xia, and C.~Shen,
  ``Twins: Revisiting the design of spatial attention in vision transformers,''
  in \emph{Advances in Neural Information Processing Systems 34: Annual
  Conference on Neural Information Processing Systems 2021, NeurIPS 2021,
  December 6-14, 2021, virtual}, 2021, pp. 9355--9366.

\bibitem{Chen2021RegionViTRA}
C.~Chen, R.~Panda, and Q.~Fan, ``Regionvit: Regional-to-local attention for
  vision transformers,'' \emph{CoRR}, vol. abs/2106.02689, 2021.

\bibitem{Lin2014Mscoco}
T.~Lin, M.~Maire, S.~J. Belongie, J.~Hays, P.~Perona, D.~Ramanan,
  P.~Doll{\'{a}}r, and C.~L. Zitnick, ``Microsoft {COCO:} common objects in
  context,'' in \emph{Computer Vision - {ECCV} 2014 - 13th European Conference,
  Zurich, Switzerland, September 6-12, 2014, Proceedings, Part {V}}, vol. 8693,
  2014, pp. 740--755.

\bibitem{Ren2022BeyondFD}
P.~Ren, C.~Li, G.~Wang, Y.~Xiao, Q.~Du, X.~Liang, and X.~Chang, ``Beyond
  fixation: Dynamic window visual transformer,'' \emph{CoRR}, vol.
  abs/2203.12856, 2022.

\bibitem{Zhou2019ade20k}
B.~Zhou, H.~Zhao, X.~Puig, T.~Xiao, S.~Fidler, A.~Barriuso, and A.~Torralba,
  ``Semantic understanding of scenes through the {ADE20K} dataset,'' \emph{Int.
  J. Comput. Vis.}, vol. 127, no.~3, pp. 302--321, 2019.

\end{thebibliography}

\end{document}